\documentclass[conference,compsoc]{IEEEtran}
\IEEEoverridecommandlockouts
\usepackage{amsmath,amssymb,amsfonts}
\usepackage{algorithmic}
\usepackage{graphicx}
\usepackage{textcomp}
\usepackage{xcolor}
\def\BibTeX{{\rm B\kern-.05em{\sc i\kern-.025em b}\kern-.08em
    T\kern-.1667em\lower.7ex\hbox{E}\kern-.125emX}}

\usepackage{times}
\usepackage{epsfig}

\PassOptionsToPackage{numbers, compress}{natbib}
\usepackage{natbib}
\usepackage[utf8]{inputenc} 
\usepackage[T1]{fontenc}    
\usepackage{url}            
\usepackage{booktabs}       
\usepackage{nicefrac}       
\usepackage{microtype}      

\makeatletter
\g@addto@macro{\UrlBreaks}{\UrlOrds}
\makeatother

\usepackage{algorithm}
\usepackage[ruled,linesnumbered,algo2e]{algorithm2e}
\DeclareMathOperator*{\argmin}{arg\,min}
\usepackage{enumitem}
\usepackage{multirow}

\usepackage{etoolbox}
\makeatletter
\patchcmd{\@makecaption}
  {\scshape}
  {}
  {}
  {}
\makeatother

\usepackage[
singlelinecheck=false 
]{caption}

\def\tablename{Table}

\makeatletter
\def\@IEEEsectpunct{.\ \,}
\def\paragraph{\@startsection{paragraph}{4}{\z@}{1.5ex plus 1.5ex minus 0.5ex}%
{0ex}{\normalfont\normalsize\sffamily\bfseries}}
\makeatother

\usepackage[breaklinks=true,bookmarks=false]{hyperref}

\begin{document}
\title{OGAN: Disrupting Deepfakes with an Adversarial Attack that Survives Training}

\author{\IEEEauthorblockN{Eran Segalis}
\IEEEauthorblockA{\textit{Independent Scholar}\\
segaliseran@gmail.com}
\and
\IEEEauthorblockN{Eran Galili}
\IEEEauthorblockA{\textit{Independent Scholar}\\
Eran.Galili.Academic@gmail.com}
}

\maketitle
\thispagestyle{plain}
\pagestyle{plain}

\begin{abstract}
Recent advances in autoencoders and generative models have given rise to effective video forgery methods, used for generating so-called ``deepfakes''. Mitigation research is mostly focused on post-factum deepfake detection and not on prevention. We complement these efforts by introducing a novel class of adversarial attacks---training-resistant attacks---which can disrupt face-swapping autoencoders whether or not its adversarial images have been included in the training set of said autoencoders. We propose the Oscillating GAN (OGAN) attack, a novel attack optimized to be training-resistant, which introduces spatial-temporal distortions to the output of face-swapping autoencoders. To implement OGAN, we construct a bilevel optimization problem, where we train a generator and a face-swapping model instance against each other. Specifically, we pair each input image with a target distortion, and feed them into a generator that produces an adversarial image. This image will exhibit the distortion when a face-swapping autoencoder is applied to it. We solve the optimization problem by training the generator and the face-swapping model simultaneously using an iterative process of alternating optimization. Next, we analyze the previously published Distorting Attack and show it is training-resistant, though it is outperformed by our suggested OGAN. Finally, we validate both attacks using a popular implementation of FaceSwap, and show that they transfer across different target models and target faces, including faces the adversarial attacks were not trained on. More broadly, these results demonstrate the existence of training-resistant adversarial attacks, potentially applicable to a wide range of domains. 
\end{abstract}
\begin{figure}[!htb]
    \includegraphics[width=1.0\columnwidth]{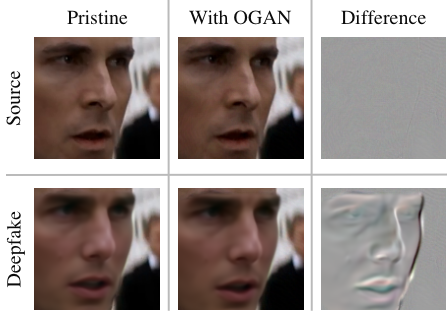}
    \caption{\textbf{OGAN visualization.} In the top row, a pristine image, followed by the OGAN-protected image, and a visualization for their difference. In the bottom row, a deepfake model applied to these two images, and the difference of its outputs. In the source data we see no observable difference, while in the deepfaked data we see a significant artifact - these artifacts create visible oscillations in videos.}
    \label{fig:teaser}   
\end{figure}

\section{Introduction}
\label{intro}
Recent improvements in deep learning have contributed to the rise of questionable applications that perform synthetic image rendering, known as deepfakes. They are used maliciously in various ways, from ``face swapping'' - replacing a persons face in a video with another to misrepresent the target \cite{deepfakes-github} to ``face reenactment'' \cite{DBLP:conf/cvpr/ThiesZSTN16} - which alters the expressions of a target person in a video by transferring the expressions of a source person to the target.

Efforts to combat such deepfake systems have mostly focused on detection \cite{DBLP:journals/corr/abs-1812-02510, DBLP:conf/biosig/KhodabakhshRRWB18,  DBLP:conf/mipr/MarraGCV18, DBLP:conf/iccv/RosslerCVRTN19, DBLP:conf/ijcai/WangJMXHWL20, DBLP:conf/cvpr/WangW0OE20, DBLP:conf/cvpr/ZhouHMD17}, rather than prevention. While identifying content as fraudulent is important, when it comes to content that besmirches a person's reputation - its mere existence might mean that the damage is already done.

Recently, a new approach to prevent deepfakes has surfaced -- implementing adversarial attacks on the malicious deepfake models. An adversarial attack on a given model involves applying minute changes to a given input that are imperceptible by a human, but should the model be applied to the modified input, its output would be erroneous. Two examples of such successful attacks against deepfake systems have been shown in \cite{DBLP:journals/corr/abs-2003-01279, DBLP:conf/wacv/YehCTW20}.

Such adversarial attacks are effective in defeating deepfakes when the model is unlikely to be trained on data that includes the adversarial samples. Unfortunately, this assumption does not always hold. For example, in a fake news scenario such as \cite{forbes-deepfakes-bad}, a deceiver might train a face-swapping model from scratch on videos of a political figure, in order to swap their face with another. Protecting the political figure's videos by injecting adversarial samples would cause these samples to be included in the deepfake model's training set, which could thwart the attack.

In this work, we propose a new family of attacks: \emph{training-resistant adversarial attacks}. These attacks are similarly applied against a given model to produce adversarial samples, but they are a stronger form of adversarial attacks since they survive training. When the attacked model is applied to these samples, it will yield an erroneous output, whether or not these adversarial samples were included in its training data. 

We demonstrate such a training-resistant attack against the face-swapping application of deepfake. We chose to focus on this application both because of the increasingly widespread usage of face-swapping in deepfakes, and because face-swapping is a good representative case of a deepfake model, and a successful attack on it could likely be generalized to more applications.

Our attack aims to inject minute perturbations to source video frames, so that when a face-swapping model is applied to them, the output includes visible spatial-temporal distortions that warps it, making the swap evident to a human eye (Figure \ref{fig:teaser}). To achieve this goal and the property of training resistance, we formulate the objective as a general bilevel optimization problem, where we train a face-swapping model instance and an adversarial sample generator against each other; we choose this path over a more specific Minimax problem to avoid modifying the \emph{FaceSwap} autoencoder's loss function, while keeping the flexibility to modify the adversarial network's loss function, and direct it toward generating samples that result in effective disruptions. 

Our main contributions are:
\begin{itemize}[itemsep=4pt,parsep=4pt]
\item We introduce the concept of \emph{training resistant adversarial attack}, motivated by real world deepfake applications where the training process of the model will likely include adversarial examples generated to defeat it.
\item We study the Distorting Attack by Yeh \emph{et~al.} \cite{DBLP:conf/wacv/YehCTW20}, and empirically show it satisfies the training resistance property.
\item We propose the \emph{Oscillating GAN (OGAN)} attack method, which improves on this previous work.
\item We evaluate OGAN and the Distorting Attack on several implementations of face-swapping, the most popular deepfake system, and show OGAN outperforms it in scenarios where the attacked model's training set includes the adversarial samples.
\end{itemize}

\section{Related Work}
\label{related_Work}
Adversarial attacks had been extensively studied in the context of classification problems \cite{DBLP:journals/access/AkhtarM18, DBLP:journals/corr/BalujaF17, DBLP:conf/sp/Carlini017, DBLP:journals/corr/GoodfellowSS14, DBLP:conf/iclr/KurakinGB17a, DBLP:conf/iclr/MadryMSTV18, DBLP:conf/cvpr/Moosavi-Dezfooli17, DBLP:journals/corr/SzegedyZSBEGF13, DBLP:conf/iclr/TramerKPGBM18}, but less research was published on their effect on generative models and autoencoders \cite{ DBLP:journals/corr/abs-1806-04646, DBLP:conf/sp/KosFS18, DBLP:journals/corr/TabacofTV16}. Tabacof \emph{et~al.} \cite{DBLP:journals/corr/abs-1806-04646, DBLP:journals/corr/TabacofTV16} and Kos \emph{et~al.} \cite{DBLP:conf/sp/KosFS18} explore adversarial attacks against Variational Autoencoders (VAE), where Autoencoders and VAE-GAN models are used for image compression.
Wang \emph{et~al.} \cite{DBLP:journals/ral/WangCY20} adapt adversarial attacks to image-to-image translation tasks under both paired and unpaired settings. Additionally for the paired setting they adapt a poisoning attack on the target domain. Yeh \emph{et~al.} \cite{DBLP:conf/wacv/YehCTW20} and Ruiz \emph{et~al.} \cite{DBLP:journals/corr/abs-2003-01279} are two concurrent works to ours, which explore adversarial attacks to disrupt deepfake models. Yeh \emph{et~al.} propose a Distorting attack in which the image translation model output becomes corrupt and a Nullifying attack in which the model becomes the identity mapping. Ruiz \emph{et~al.} \cite{DBLP:journals/corr/abs-2003-01279} explore distorting attacks and extend them to conditional image translation networks. Additionally they adapt adversarial training \cite{DBLP:conf/iclr/MadryMSTV18} for conditional image translation GANs. Willetts \emph{et~al.} \cite{willetts2019improving} explore defenses against adversarial attacks for VAE.

All of these attacks assume that the target models were trained using pristine data, a reasonable assumption for many applications. Unfortunately, for many other deepfake tasks this assumption doesn't hold. For example, in a face-swapping scenario, a deceiver aims to swap the faces of \(A\) and \(B\) for some video \(v_A\), which requires training a specific \(A \rightarrow B\) model, and therefore collecting the corresponding training data. Since, in this scenario, \(v_A\) is available to the deceiver, the training data is likely to include images from \(v_A\) - which is the video we aim to protect, and therefore, will contain our adversarial images. 
Such cases may pose a challenge to attacks proposed by earlier works, which did not study the effect of the target model training on the adversarial images.

Consequently, an effective attack on this scenario should assume adversarial images might be included in the training data so the attack must be training-resistant, as defined in section~\ref{intro}. On the other hand, unlike in a poisoning attack (where one aims to poison the training data with bad inputs), our training-resistant attack should also assume that they might not be there, and succeed either way. This independence makes our attack more robust. 

In this work, we first study the distorting attack by Yeh \emph{et~al.} \cite{DBLP:conf/wacv/YehCTW20}, and show that it has the training-resistant property. In addition, we propose a new attack, OGAN, which has been optimized specifically for training resistance - and finally, we show that OGAN outperforms the distorting attack in cases where training resistance is required (defeating deepfakes). To the best of our knowledge, we are the first to introduce and study the notion of training-resistant adversarial attacks in any domain.

\begin{figure*}[!htb]
    \centering
    \includegraphics[width=2.0\columnwidth]{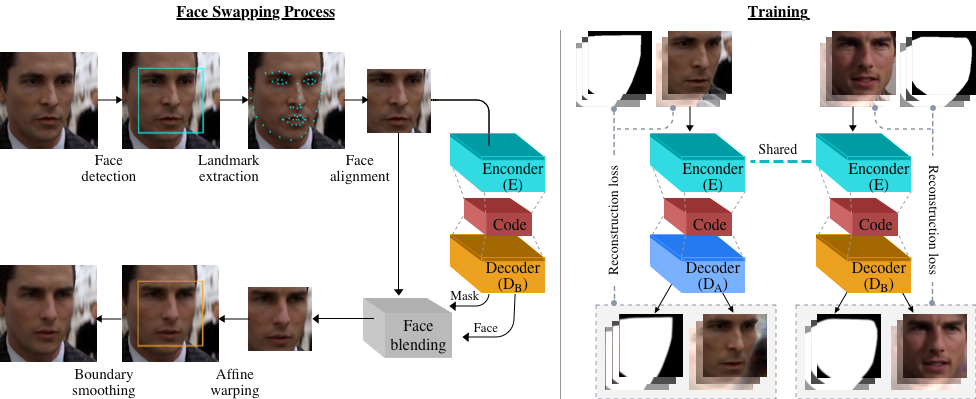}
    \caption{\textbf{Face swapping overview:} The application and training processes of the face swapping process. For further details, see section \ref{DeepFake_generation}.}
    \label{fig:faceswap_process}
\end{figure*}
\section{Method}
\label{method}
\subsection{DeepFake generation}
\label{DeepFake_generation}
While the term deepfake has become synonymous with its result of replacing a face in a video, it also refers to a specific face swapping method. The most notable implementation of this method is FaceSwap \cite{deepfakes-github}, as has been analyzed by \cite{DBLP:conf/iccv/RosslerCVRTN19} and many others. In this section, we will briefly describe this implementation of deepfake, an overview of which is shown in Figure \ref{fig:faceswap_process}.

This system receives as input an image sequence featuring source face A (a video, in this context, is viewed as an image sequence), and a sequence featuring target face B. First, an extraction phase extracts the faces from the images, aligns them, and creates masks that indicate where each face is located in each image. Next, two autoencoders with a shared encoder are trained to reconstruct images of the source and target faces. These autoencoders are trained on an augmentation of the input images, created via rotations, translations, magnifications and minute random color changes. Then, they are optimized using a reconstruction loss function, which consists of a mask loss function and a face loss function which is weighted by the input mask. The reconstruction loss function is thus defined as follows:
\begin{align}
\mathcal{L}_{rec}(x^{face},x^{mask},y,m) = &\quad\ \mathcal{L}_{face}(x^{face},y,m)   \\
                                           & + \mathcal{L}_{mask}(x^{mask},m)  \nonumber
\end{align}
\begin{align}
&\mathcal{L}_{face}(x^{face},y,m) = \| \big(x^{face} - y \big) \odot m \|_{1}\\
&\mathcal{L}_{mask}(x^{mask},m) = \| x^{mask} - m \|_{1}
\end{align}
where \(x^{face}\) and \(x^{mask}\) are the predicted face and mask, \(y\) is the input image, \(m\) is the input mask and \(\odot\) denotes a point-wise multiplication. We will refer to this loss function as the FaceSwap loss.

This yields an encoder and decoder pair for both face A and face B, where the decoders output both an image and face mask.

Finally, a swapped face is produced by applying the trained encoder and decoder of face B to the target images of face A. The output face is blended into the target image using the face mask. 

\paragraph{DeepFake architectures} \label{DeepFake-architectures}
FaceSwap \cite{deepfakes-github} offers several autoencoder architectures and configurations, but training a FaceSwap model is an expensive task that requires several days on high-end GPUs. Hence, a full evaluation of all the models is beyond the scope of this paper. 
Accordingly, we chose 3 architectures for our research:

\begin{itemize}[itemsep=4pt,parsep=4pt]
\item \textbf{realface}: This architecture uses skip connections for both its encoder and its decoders. It also uses an unbalanced framework where the autoencoder for the target face \(B\) has additional layers.
\item \textbf{dfl-h128}: DeepFaceLab \cite{DeepFaceLab-github} is the one of the most popular deepfake implementations. This model includes a \(128\times128\) pixels input model without skip connections, and closely reassembles the original model implemented in FaceSwap \cite{deepfakes-github}.
\item \textbf{dfl-sae}: Another DeepFaceLab \cite{DeepFaceLab-github} architecture, this one uses skip connections only for the decoders.
\end{itemize}
These models provide us with diverse architecture types and represent the two most popular implementations of deepfake - \cite{deepfakes-github, DeepFaceLab-github}.

\subsection{Optimization problem formulation}
Our system's objective is to add a tamper-evident feature to videos, disrupting attempts to manipulate them using deepfake. Thus, for a given video \(v\) including the face \(A\), our system will output a modified video \(\bar{v}\), where the differences between \(v\) and \(\bar{v}\) are imperceptible to a human observer. However, when the deepfake system is executed on \(\bar{v}\) with some target face \(B\), the resulting \(f_B(\bar{v})\) will include, instead of a seamless replacement of \(A\) with \(B\), major human-visible artifacts identifying the deepfake tampering performed - thus defeating the attempted face swapping.

To achieve this goal, we use a class of disruptions to modify each frame in video \(v\) which are imperceptible in \(\bar{v}\), but cause a change in the location, scaling and angle of face \(B\) in \(f_B(\bar{v})\) - a change we represent using an affine transformation, N.

To formulate this, we define an adversarial generator \(G(x,N)\) which receives as input a face image \(x\) and the target affine transformation \(N\). \(G(x,N)\) outputs a modified face image satisfying:
\begin{align}
G(x,N) = \argmin_{y} \quad & \mathcal{L}_{adv}(y,x,N,m) \\
\textrm{s.t.} \quad & \|y - x\|_{\infty} \leq \varepsilon \label{eq:l_inf_constraint}
\end{align}
Where \(\varepsilon\) is used to control the magnitude of the adversarial perturbation which is common in adversarial settings \cite{DBLP:journals/corr/GoodfellowSS14}, and \(\mathcal{L}_{adv}\) is the adversarial loss function defined by:
\begin{equation}
\mathcal{L}_{adv}(x_{adv},x,N,m) = \mathcal{L}_{face}(f_{A}^{face}(x_{adv}), N(x), m)
\end{equation}
Where \(x\) and \(m\) are the original face image and mask, \(x_{adv}\) is the pertubated face image, \(f_{A}^{face}\) is the face output of the autoencoder \(f_A\)  for face \(A\) and \(N\) is the target affine transformation.

For the adversarial loss, we use the autoencoder \(f_A\) of face \(A\). We do this to keep the adversarial loss an internal process, i.e. defined by face \(A\) as much as possible, since an internal process will likely increase the odds for transferability of our adversarial attack to target faces other than \(B\).

Now, let \(\mathcal{D}_A,\mathcal{D}_B\) be the datasets used for training the FaceSwap autoencoders for faces \(A\) and \(B\). Let \(\mathcal{P}_A \subseteq \mathcal{D}_A\) be the subset of data we can control and would like to protect. For each \((x, m_x) \in \mathcal{P}_A\) we pick a distortion transformation \(N_{x}\). We aim to find an adversarial generator such that:
\begin{align}
    G^{\star} = & \argmin_{G}  \sum_{(x,m_x) \in \mathcal{P}_A} \mathcal{L}_{adv}(G(x,N_x),x,N_x,m_x) \\
    &\ \textrm{s.t.} \quad (f_{A},f_{B}) \in \bigg\{ \argmin_{f_{A},f_{B}}
    \  \mathcal{L}_{B}(\mathcal{D}_B) + \mathcal{L}_{A}(\mathcal{D^{\prime}}_{A})
     \bigg\} \nonumber
\end{align}
Where \(\mathcal{L}_{A}\) and \(\mathcal{L}_{B}\) are the FaceSwap losses for faces \(A\) and \(B\) as defined by:
\begin{equation}
\mathcal{L}_{A}(\mathcal{D}) = \sum_{(x,m_x) \in \mathcal{D}} \mathcal{L}_{rec}(f_{A}^{face}(x),f_{A}^{mask}(x),x,m_x)
\end{equation}
\begin{equation}
\mathcal{L}_{B}(\mathcal{D}) = \sum_{(x,m_x) \in \mathcal{D}} \mathcal{L}_{rec}(f_{B}^{face}(x),f_{B}^{mask}(x),x,m_x)
\end{equation}

Where \(f_{A}^{face},f_{A}^{mask},f_{B}^{face},f_{B}^{mask}\) are the face and mask outputs from the autoencoders \(f_{A}\)  and \(f_B\) of faces \(A\) and \(B\) accordingly. \(\mathcal{D^{\prime}}_{A}\) consists of the images from \(\mathcal{P}_A\) perturbated by \(G\) and the rest of the images of \(\mathcal{D}_A\). More formally:
\begin{equation}
\mathcal{D^{\prime}}_{A} =  \big\{(G(x,N_x),m_x)\ \big|\ (x,m_x) \in \mathcal{P}_A\big\} \cup (\mathcal{D}_A \setminus \mathcal{P}_A)
\end{equation}
\paragraph{Remark}
An important factor in our problem formulation is the exclusion of the \(\ell_{\infty}\) constraint. This is due to our design of \(G\), in which we enforce the \(\ell_{\infty}\) constraint of Equation  \ref{eq:l_inf_constraint} via the network architecture itself.

\begin{figure*}[!tb]
    \centering
    \captionsetup{justification=centering}
    \includegraphics[width=1.7\columnwidth]{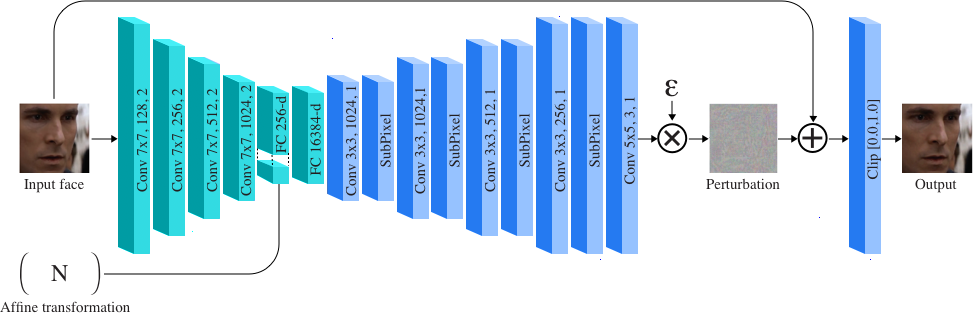}
    \caption{\textbf{OGAN generator architecture.}}
    \label{fig:generator_architecture}
\end{figure*}
\subsection{Training procedure}
\begin{algorithm2e}[!tb] 
    \caption{OGAN training algorithm}
    \label{alg:training-algo}
    \For{\(iter=1\) {\bfseries to} \(T\)}
    {
    Train autoencoders \(f_A\) and \(f_B\) on a single batch
    }
    \For{\(p\) {\bfseries in} \(\mathcal{P}_A\)}
    {
     \(N_p \leftarrow\)  Sample random affine transormation 
    }
    \For{\(epoch=1\) {\bfseries to} \(M\)}
    {
        \For{\(i=1\) {\bfseries to} number of batches in \(\mathcal{P}_A\)}
        {
            \(b_p,b_m,b_N \leftarrow \mathcal{P}_A\)-batch: images, masks and transformations\\
            \For{\(i=1\) {\bfseries to} \(BatchSize\)}
            {
                \(b^{\prime}_{p,i} \leftarrow b_{N,i}(b_{p,i}) \) // Transform face
            }
            \For{\(k=1\) {\bfseries to} \(BatchIters\)}
            {
                // Train \(G\) on batch \\
                \(G \leftarrow opt_{G}(b_p, b_N, b^{\prime}_p)\)
            }
            \(b^{adv}_p \leftarrow G(b_p, b_N) \) \\
            \(b^{aug}_p, b^{aug}_m \leftarrow Augment(b^{adv}_p, b_m)\) \\
            \(f_A \leftarrow opt_{f_A}(b^{aug}_p, b^{aug}_m)\) \\
            \texttt{\\} 
            \(c^{aug}_p, c^{aug}_m  \leftarrow\) Augmented \(\mathcal{D}_B\)-batch\\
            \(f_B \leftarrow opt_{f_B}(c^{aug}_p, c^{aug}_m)\) 
        }
        \For{\(i=1\) {\bfseries to} number of batches in \((\mathcal{D}_A \setminus \mathcal{P}_A)\)}
        {
            \(b^{aug}_p, b^{aug}_m  \leftarrow\) Augmented \(\mathcal{D}_A\)-batch\\
            \(f_A \leftarrow opt_{f_A}(b^{aug}_p, b^{aug}_m)\) \\
             \texttt{\\} 
            \(c^{aug}_p, c^{aug}_m  \leftarrow\) Augmented \(\mathcal{D}_B\)-batch\\
            \(f_B \leftarrow opt_{f_B}(c^{aug}_p, c^{aug}_m)\) 
        }
    }
    \Return \(G,\{N_p\}\)
\end{algorithm2e}
Next, we proceed to the design of the training algorithm. We begin by training a FaceSwap model for \(T\)
iterations, resulting in the autoencoders \(f_A\) and \(f_B\) for faces \(A\) and \(B\) respectively.

We continue by selecting a family of affine transformations \(\{N_p\}\), for the target disruptions we plan on inducing. We first define the parameters \(\Theta\)  and \(\Psi\) , and then for each image \(p\), we uniformly sample from \([-\Theta,\Theta]\) and \([-\Psi,\Psi]\times [-\Psi,\Psi]\)  a rotation angle \(\theta\) and shift \((\psi_x,\psi_y)\) respectively, and define \(N_p\) to be the affine transformation operating on \(p\) by rotating it by \(\theta\), then shifting it by \((\psi_x,\psi_y)\).

For the next part of the training algorithm, we note that our bilevel optimization problem is closely related to the one used in poisoning attacks such as \cite{DBLP:conf/ccs/Munoz-GonzalezB17}. Therefore, we similarly solve the optimization problem by running an iterative process of alternating optimization of both \(G\) (via the adversarial loss \(\mathcal{L}_{adv}\)) and \(f_A,f_B\) (via the FaceSwap losses \(\mathcal{L}_{A}, \mathcal{L}_{B}\)), as a starting point for this proccess we use a random initilization for our generator \(G\) and for the FaceSwap autoencoders  \(f_A,f_B\) we use autoencoders obtained from training a FaceSwap model for \(T\)  iterations.

Finally, in order to stabilize the training process and to achieve stronger distortions of the output, we allow our generator extra training cycles for each batch we train on. The resulting algorithm is summarized in Algorithm~\ref{alg:training-algo}.

\paragraph{Network architecture} For our adversarial generator, we choose to rely on the work in \cite{DBLP:journals/corr/BalujaF17, DBLP:conf/nips/FengCZ19} and use an autoencoder architecture. Specifically, we adapted the dfl-h128 autoencoder used in FaceSwap \cite{deepfakes-github}, which we've chosen since it is the model we train against in Algorithm~\ref{alg:training-algo}. To address the \(\ell_{\infty}\) constraint, we use \(tanh\) non-linearity in our last convolutional layer and multiply the result by \(\varepsilon\). A detailed description of the generator network's architecture is shown in figure \ref{fig:generator_architecture}.

\subsection{Distorting attack}
Yeh \emph{et~al.} \cite{DBLP:conf/wacv/YehCTW20} and Ruiz \emph{et~al.} \cite{DBLP:journals/corr/abs-2003-01279} have proposed an adversarial attack on image translation models (a class that includes FaceSwap). This Distorting Attack defines its adversarial loss as a function that aims to move a target model's output for an adversarial example as far as possible from its output for the original example. 
Next, the attack utilizes Projected Gradient Descent (PGD) \cite{DBLP:conf/iclr/MadryMSTV18} to generate the adversarial samples (the protected images) from the original samples (the input images). More formally, let \(x\) be an input image, let \(H\) be an image translation model, let \(\varepsilon\) be a perturbation bounding parameter and \(\alpha\) a learning rate parameter. PGD consists of a random initialization and an update rule:
\begin{align}
x^{adv}_{0} &= x + noise\\ \nonumber
x^{adv}_{t+1} &= \text{clip}_{x,\varepsilon}(x^{adv}_{t} - \alpha \cdot \text{sign}(\nabla_{x^{adv}} \mathcal{L}(x^{adv}_{t},x))
\end{align}
Where \(x^{adv}_{t}\) is the adverserial example at the \(t_{th}\) iterarion, noise is a random vector of the same dimensions as x and whose elements are in \([-\varepsilon,\varepsilon]\), and \(\text{clip}_{x,\varepsilon}(z)\) denotes element-wise clipping \(z\), with \(z_{i,j}\) clipped to the range \([x_{i,j} - \varepsilon, x_{i,j} + \varepsilon]\) and then clipped to valid image range and \(\mathcal{L}\) the adverserial loss used for the attack.

In the Distorting attack the adverserial loss \(\mathcal{L}\) equals the distortion loss \(\mathcal{L}_{dist}\) defined by:
\begin{equation}
\mathcal{L}_{dist}(x^{adv},x)  = -\| H(x^{adv}) - H(x) \|_{2}
\end{equation}

For our attack setting, \(H(x)\) is the deepfaked image produced from \(x\) using the \(f_{B}\) autoenconder, specifically:
\begin{equation}
H(x) = f_{B}^{face}(x) \odot  f_{B}^{mask}(x)  + x \odot (1 - f_{B}^{mask}(x))
\end{equation}
Where \(\odot\) denotes a point-wise multiplication and \(f_{B}^{face},f_{B}^{mask}\) are the face and mask outputs from the autoencoder \(f_B\) of face \(B\).

\section{Experiments}
\label{experiments}
We evaluate our attack against the face-swapping models described in section \ref{DeepFake-architectures}, with some slight adjustments so the models would have the same input size and they could be trained with the same batch size while fitting in our GPU's memory. First, we match the input size of the realface model to \(128 \times 128\). Next, we slightly decrease the sizes of realface (by setting dense\_nodes to \(1408\)) and dfl-h128 (by setting low\_memory=True). This decrease in size should not affect the efficacy of our technique on the original models.

\subsection{Datasets}
\label{datasets}
We collected several videos from YouTube for 3 people \(A,B,C\). For each person, we first extracted their face images from the videos using S\(^3\)FD \cite{DBLP:journals/corr/abs-1708-05237} and aligned them using FAN \cite{DBLP:conf/iccv/BulatT17}. Then, we removed blurry images and the face images of other people appearing in the videos. This resulted in a set of about \(5000\) face images for each person; these are the datasets we will train FaceSwap models on, \(\mathcal{D}_A\), \(\mathcal{D}_B\) and \(\mathcal{D}_C\). For the dataset of data we control, \(\mathcal{P}_A\), we arbitrarily choose person \(A\) and focus on the subset of \(444\) of face \(A\)'s images extracted from one of their videos, \(v_A\). The remainder of face \(A\)'s images, \(\mathcal{D}_A \setminus \mathcal{P}_A\), emulates data a faceswapping deceiver might have access to, but which we do not control.

\subsection{Generation of adversarial images}
To generate the adversarial images from both OGAN and the Distorting Attack, we begin by training a FaceSwap model with the dfl-h128 architecture using the datasets \(\mathcal{D}_A\) and \(\mathcal{D}_B\) for \(T=200000\) iterations. The model is optimized with the Adam optimizer, configured with the following hyperparameters: \(\beta_1=0.5,\ \beta_2=0.999\), a learning rate of \(5 \times 10^{-5}\) and a batch size of \(64\).

Next, for the Distorting Attack, we use the autoencoder \(f_B\) from this trained model to generate the adversarial images for the original face images in \(\mathcal{P}_A\) over \(1000\) iterations, with the hyperparameters: \(\varepsilon=\frac{2}{255},\ \alpha=0.001\). This results in the adversarial image set we mark as \(O^{Dist}_{A}\).

Finally, for OGAN, we train our generator against the trained FaceSwap model using Algorithm~\ref{alg:training-algo}, with the FaceSwap model we've already trained, and the following hyperparameters:  \(\varepsilon=\frac{2}{255},\ \Theta=10,\ \Psi=12.8,\ BatchSize=64,\  BatchIters=8\).
The OGAN networks are also trained using the Adam optimizer with the following hyperparameters: \(\beta_1=0.9,\ \beta_2=0.9999\), a learning rate of \(5 \times 10^{-5}\) and a batch size of \(64\). The generator has been trained for \(M=1024\) epochs using \(\mathcal{D}_A, \mathcal{P}_A\) and \(\mathcal{D}_B\). As explained in Algorithm~\ref{alg:training-algo} above, this results in the trained adversarial generator, as well as the output affine transformations. We use these generator and transformations to calculate the adversarial images of the face images in \(\mathcal{P}_A\), resulting in the adversarial image set  \(O^{OGAN}_{A}\).

Finally, we will patch the faces from both \(O^{OGAN}_{A}\) and \(O^{Dist}_{A}\) back onto their original video \(v_A\). We do this so we are later able to simulate an end-to-end application of deepfake techniques on the videos. After patching the face images from each of  \(O^{OGAN}_{A}\) and \(O^{Dist}_{A}\)  onto \(v_A\), we now have the adversarial videos (which are the protected videos) \(\bar{v}^{OGAN}_{A}\) and  \(\bar{v}^{Dist}_{A}\) respectively. 

\begin{figure*}[!htb]
    \centering
    \includegraphics[width=1.7\columnwidth]{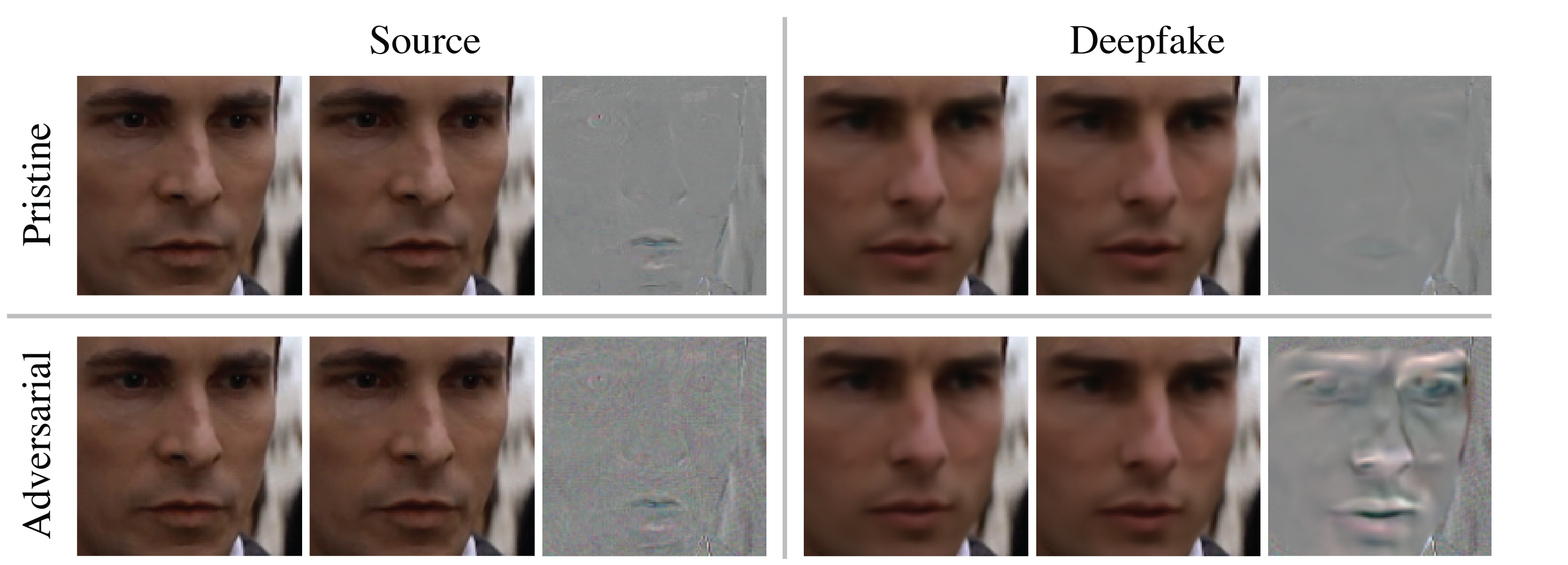}
    \caption{\textbf{Perturbation visualization.} Samples from a dfl-h128 face swapping model trained for \(500000\) iterations on \(\mathcal{D}^{OGAN}_A\). First row from left to right: Pristine consecutive face images from our video, visualization of their difference, their face swapping outputs and a visualization of their difference. Second row from left to right: adversarial perturbations of the same consecutive images, visualization of their difference, their face swapping outputs and a visualization of their difference. This shows that the difference created by adding the adversarial perturbations to the pristine video resembles white noise and is thus not noticeable, but applying face swapping results in major, noticeable artifacts. }
    \label{fig:temporal_demo_500001}
\end{figure*}
\subsection{Metrics}
\paragraph{Spatial-temporal score}
To quantitatively measure video distortion, and to evaluate OGAN and the Distorting Attacks, we use the temporal consistency metric \(E_{tmp}\) \cite{DBLP:conf/cvpr/HuangWLMJZLL17, DBLP:conf/cvpr/ChengCC20}. This metric quantifies the overall differences between consecutive frames in a video by averaging the pixel-wise Euclidean difference in color between frames, as defined by:
\begin{equation}
E_{tmp}(y) = \gamma \sqrt{\frac{1}{d(n-1)}\sum_{i=1}^{n-1} \|y_{i+1} - y_{i} \|^{2}_{2}}
\end{equation}
where \(y\) is a sequance of video frames, \(y_{i}\) is the \(i\)-th frame, \(n\) is number of frames in the video, \(d\) is the number of pixels in each frame and \(\gamma\) is a scaling constant, set to \(10^3\) \cite{DBLP:conf/cvpr/ChengCC20}.

Next, we note that consecutive video frames in a non-static video (meaning, a video which is not a sequence of identical images) would exhibit a natural spatial-temporal difference. Thus, in addition to \(E_{tmp}\) we compute \(S_{tmp}\), which is the normalized \(E_{tmp}\) metric, to obtain a score that's independent of the specific video.
\(S_{tmp}\) is computed by:
\begin{equation}
S_{tmp}(\bar{v}_{A \rightarrow K}, v_{A \rightarrow K}) = \frac{E_{tmp}(\bar{v}_{A \rightarrow K})}{E_{tmp}(v_{A \rightarrow K})} - 1
\end{equation}
Where K is a target face (either face \(B\) or face \(C\)), \(v_{A \rightarrow K}\) is a deepfaked video of \(v_{A}\) (i.e. where we swapped face \(A\) to face \(K\)) and \(\bar{v}_{A \rightarrow K}\) is deepfaked video of \(\bar{v}_{A}\) (where \(\bar{v}_{A}\) is the adversarial video obtained from \(v_{A}\)).

Since the ratio \(\frac{E_{tmp}(\bar{v}_{A \rightarrow K})}{E_{tmp}(v_{A \rightarrow K})}\) is \(1\) if the adversarial attack did not impact the deepfake model's output at all, we adjust \(S_{tmp}\)'s range by substracting 1 from the result.




\subsection{Attacks evaluation}
As before (section \ref{datasets}), we extract face images from \(\bar{v}^{OGAN}_{A}\) and \(\bar{v}^{Dist}_{A}\) using S\(^3\)FD and FAN. We then combine each of these face image sets with the face images from \(\mathcal{D}_A \setminus \mathcal{P}_A\) - resulting in the new datasets \(\mathcal{D}^{OGAN}_A\) and \(\mathcal{D}^{Dist}_A\) respectively.

\paragraph{Training resistance}
We first evaluate the attacks against face swapping models trained using the datasets\(\mathcal{D}^{OGAN}_A\) and \(\mathcal{D}^{Dist}_A\), which include our adversarial images. To do this, we train a dfl-h128 model for the task of face swapping \(A \rightarrow B\) for each of these datasets. These models (and other face swapping models mentioned later in this section) were trained for  \(200000\) iterations using the Adam optimizer with the following hyperparameters: \(\beta_1=0.5,\ \beta_2=0.999\), a learning rate of \(5 \times 10^{-5}\) and a batch size of \(64\). The training was executed using the FaceSwap code \cite{deepfakes-github} using its default hyperparameters, and the same number of training iterations as were used to generate the FaceForensics dataset \cite{DBLP:conf/iccv/RosslerCVRTN19}, which is used for the detection of face swapped videos.

Figure~\ref{fig:temporal_demo_500001} shows that injecting the OGAN adversarial perturbations to the pristine video creates a difference between consecutive frames that's similar to white noise added to the natural difference between the matching pristine frames - this is why the perturbations are not observable in the adversarial source video. However, the difference between consecutive frames of this video after a deepfake model is applied to it resembles a face - this is because the attack caused the face's location to shift. This shift is
noticeable in the disrupted video, as can be seen in the video in the supplementary material. Perturbations injected by the Distorting Attack create a similar effect\footnote{See the first demo video in \url{https://youtu.be/_XEcCtC7EEw}}.

Next, we analyze the attacks perfomance, on the dfl-h128 FaceSwap models we trained, using \(S_{tmp}\) - see Table \ref{train-with-adversarial-base-model-table}. These results show that even though the face swapping models are familiar with the adversarial images (i.e., their training data includes the adversarial images), both attacks successfully created major artifacts in the resulting video. However, OGAN's impact on \(S_{tmp}\) is significantly greater.

\paragraph{Attack transferability}
Next, we analyze the transferrability of the attacks across faceswapping architectures. For this, we train both the dfl-sae and realface models for \(A \rightarrow B\) for each attack. Additionally, we train another dfl-h128 model for \(A \rightarrow C\) so we can evaluate the transferability of the attacks across different target faces.

The results of applying OGAN and the Distorting Attack to these additional models are shown in Table \ref{train-with-adversarial-transferability-table}. These results show that both attacks are applicable for different faceswap mode architectures, and for faces that the attacks were not trained on (i.e. for models swapping \(A \rightarrow Z\) for some face \(Z\)). However, OGAN outperforms the Distorting Attack on both dfl-h128 and dfl-sae. For realface, both attacks achieve relatively similar \(S_{tmp}\) scores, with a slight advantage for the Distorting Attack.

\begin{table}[!tb]
  \footnotesize
  \begin{center}
  \begin{tabular}{lcc}
    \toprule
    Method                                          & \(S_{tmp}\)      & \(E_{tmp}\) \\
    \midrule
    Distorting ’2020 \cite{DBLP:conf/wacv/YehCTW20}   & 0.019            & 46.265 \\
    OGAN (Ours)                                     & \textbf{0.061}   & 48.212 \\
    \bottomrule
  \end{tabular}
  \end{center}
  \caption{\textbf{Training resistance.} Spatial-temporal scores for attacked dfl-h128 face-swapping models, trained for \(200000\) iterations on the \(\mathcal{D}^{OGAN}_A\) and \(\mathcal{D}^{Dist}_A\) datasets (which includes our adversarial images). The \(S_{tmp}\) scores shows both attacks are training resistant, with OGAN performing significantly better.}
  \label{train-with-adversarial-base-model-table}
\end{table}
\begin{table}[!tb]
  \footnotesize
  \begin{center}
  \begin{tabular}{lcccc}
    \toprule
    \multirow{2}{*}{Method}                                             & Attacked  & Swapping &  \multicolumn{2}{c}{Temporal scores:}\\
    & model& direction & \(S_{tmp}\)       & \(E_{tmp}\) \\
    \midrule
    \multirow{3}{*}{Distorting ’2020 \cite{DBLP:conf/wacv/YehCTW20}}    & dfl-h128      & \(A \rightarrow C\)   & 0.006             & 44.262 \\
                                                                        & dfl-sae       & \(A \rightarrow B\)   & 0.059             & 48.160 \\
                                                                        & realface      & \(A \rightarrow B\)   & \textbf{0.159}    & 52.701 \\
    \midrule
    \multirow{3}{*}{OGAN (Ours)}                                        & dfl-h128      & \(A \rightarrow C\)   & \textbf{0.018}    & 44.731 \\
                                                                        & dfl-sae       & \(A \rightarrow B\)   & \textbf{0.075}    & 48.892 \\
                                                                        & realface      & \(A \rightarrow B\)   & 0.149             & 51.977 \\
    \bottomrule
  \end{tabular}
  \end{center}
  \caption{\textbf{Transferability.} Spatial-temporal scores for attacks on face-swapping models of various architectures and target faces, trained for \(200000\) iterations on the \(\mathcal{D}^{OGAN}_A\) and \(\mathcal{D}^{Dist}_A\) datasets (which include our adversarial images). These results show OGAN achieves comparable or better results to the Distorting Attack on additional face swapping architectures, and on target faces the adversarial attack wasn't trained on. }
  \label{train-with-adversarial-transferability-table}
\end{table}
\begin{table}[!tb]
  \footnotesize
  \begin{center}
  \begin{tabular}{lcccc}
    \toprule
    \multirow{2}{*}{Method}                                             & Attacked  & Swapping &  \multicolumn{2}{c}{Temporal scores:}\\
    & model& direction & \(S_{tmp}\)       & \(E_{tmp}\) \\
    \midrule
    \multirow{4}{*}{Distorting ’2020 \cite{DBLP:conf/wacv/YehCTW20}}    & dfl-h128      & \(A \rightarrow B\)   & \textbf{0.270}    & 57.289 \\
                                                                        & dfl-h128      & \(A \rightarrow C\)   & \textbf{0.078}    & 44.993 \\
                                                                        & dfl-sae       & \(A \rightarrow B\)   & \textbf{0.232}    & 55.495 \\
                                                                        & realface      & \(A \rightarrow B\)   & \textbf{0.359}    & 61.706 \\
    \midrule
    \multirow{4}{*}{OGAN (Ours)}                                        & dfl-h128      & \(A \rightarrow B\)   & 0.184             & 53.562 \\
                                                                        & dfl-h128      & \(A \rightarrow C\)   & 0.058             & 44.226 \\
                                                                        & dfl-sae       & \(A \rightarrow B\)   & 0.169             & 52.765 \\
                                                                        & realface      & \(A \rightarrow B\)   & 0.280             & 58.310 \\
    \bottomrule
  \end{tabular}
  \end{center}
  \caption{\textbf{Exclusion of adversarial images.} Spatial-temporal scores for attacks on face-swapping models that were trained for \(200000\) iterations on \(\mathcal{D}_A \setminus \mathcal{P}_A\), which does not contain any adversarial images. These results show that while OGAN was designed for the training resistant case and underperforms here, it still succeeds in achieving a large \(S_{tmp}\) score - despite the attack having no control on any part of the target model's training set.}
  \label{train-pristine-table}
\end{table}
\begin{figure*}[!htb]
    \centering
    \includegraphics[width=1.7\columnwidth]{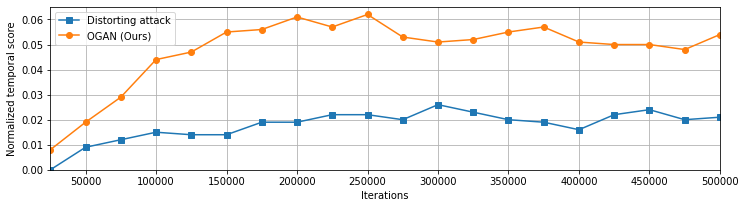}
    \caption{\textbf{Effect of extra training.} \(S_{tmp}\) scores for attacks on dfl-h128 face swapping models trained for \(500000\) iterations on the \(\mathcal{D}^{OGAN}_A\) and \(\mathcal{D}^{Dist}_A\) datasets, which contain the adversarial examples. This data shows that the \(S_{tmp}\) scores stabilize at around \(300000\) iterations, and the attacks remain successful even on a model trained for a larger number of iterations on data that includes the adversarial examples.  }
    \label{fig:plot}
\end{figure*}
\paragraph{Exclusion of adversarial images}
Next, we validate that the attacks succeed even when their adversarial images are excluded from the face swapping model's training set. While in this case, it is easier for an adversarial attack to succeed, success here will confirm that the attacks are training-resistant attacks, rather than poisoning attacks which requires injecting samples into the training process. We train dfl-h128, dfl-sae and realface on \(\mathcal{D}_A \setminus \mathcal{P}_A\) and \(\mathcal{D}_B\) for the task of \(A \rightarrow B\), and another dfl-h128 model for the task of \(A \rightarrow C\). Training is performed in the same way as before, and \(S_{tmp}\) and \(E_{tmp}\) are calculated for each attack as applied to each model.
As shown in Table \ref{train-pristine-table}, the Distorting Attack performs better than OGAN in this case. This is expected, since OGAN has been designed to focus on cases where the adversarial images are included. However, as reflected by the \(S_{tmp}\) scores, both attacks strongly outperform their results in the training resistant setting, as seen in Tables \ref{train-with-adversarial-base-model-table} and \ref{train-with-adversarial-transferability-table} and also in the greater magnitude of distortions in the second demo video\footnote{See the second demo video in \url{https://youtu.be/4dkwqWVZK0M}}.

Therefore we consider both attacks successful in this case.
\paragraph{Effect of extra training}
Finally, we investigate the effect of training the face swapping models for additional iterations on \(\mathcal{D}^{OGAN}_A\) and \(\mathcal{D}^{Dist}_A\). We train our dfl-h128 models on the task of \(A \rightarrow B\) for an additional \(300000\) iterations, and calculate \(S_{tmp}\) every \(25000\) iterations. As shown in Figure \ref{fig:plot}, the \(S_{tmp}\) score begins to stabilize after \(300000\) iterations of training, and even after \(500000\) iterations, \(S_{tmp} > 0\), with a value of \(0.021\) for the Distorting attack and a better value of \(0.054\) for OGAN. These distortions are still highly noticeable, as can be be observed in Figure \ref{fig:temporal_demo_500001}. This result confirms that even a resourceful deceiver, capable of training for many iterations, will fail in manipulating \(\bar{v}^{OGAN}_{A}\) and \(\bar{v}^{Dist}_{A}\) due to the attacks.

\section{Conclusions}
\label{conclusions}
In this work we introduced training-resistant adversarial attacks. These attacks generate adversarial samples against a given model such that when the model is applied to these samples, its intended effect is disrupted - even if the same adversarial samples were part of the model's training data. This is a key difference between such attacks and poisoning attacks, since poisoning attacks require that the adversarial samples be part of the training data.

Additionally, we developed OGAN - an attack designed for training resistance, against face swapping autoencoders. We empirically showed that the Distorting Attack by Yeh \emph{et~al.} \cite{DBLP:conf/wacv/YehCTW20} also fulfills the training resistance property, and applied both this attack and OGAN end-to-end against commonly used FaceSwap implementations.

Our results show that both attacks transfer across different face swapping models and target faces, including target faces the adversarial attacks were not trained on. Our results also show that OGAN outperforms the Distorting Attack in the training resistant case - when the target model's training data includes the adversarial samples, and that OGAN is also successful in the regular case - when the training data does not include the adversarial samples.

These results demonstrate the existence and feasibility of training-resistant adversarial attacks, potentially applicable to a wide range of domains.

\section*{Acknowledgements}
The authors would like to thank Jonathan Heimann, Roy Iarchy and Orgad Keller for many fruitful discussions and comments.

\bibliographystyle{plainnat}
\bibliography{references}

\begin{thebibliography}{33}
\providecommand{\natexlab}[1]{#1}
\providecommand{\url}[1]{\texttt{#1}}
\expandafter\ifx\csname urlstyle\endcsname\relax
  \providecommand{\doi}[1]{doi: #1}\else
  \providecommand{\doi}{doi: \begingroup \urlstyle{rm}\Url}\fi

\bibitem[Akhtar and Mian(2018)]{DBLP:journals/access/AkhtarM18}
Naveed Akhtar and Ajmal~S. Mian.
\newblock Threat of adversarial attacks on deep learning in computer vision:
  {A} survey.
\newblock \emph{{IEEE} Access}, 6:\penalty0 14410--14430, 2018.

\bibitem[Baluja and Fischer(2017)]{DBLP:journals/corr/BalujaF17}
Shumeet Baluja and Ian Fischer.
\newblock Adversarial transformation networks: Learning to generate adversarial
  examples.
\newblock \emph{CoRR}, abs/1703.09387, 2017.

\bibitem[Bulat and Tzimiropoulos(2017)]{DBLP:conf/iccv/BulatT17}
Adrian Bulat and Georgios Tzimiropoulos.
\newblock How far are we from solving the 2d {\&} 3d face alignment problem?
  (and a dataset of 230, 000 3d facial landmarks).
\newblock In \emph{{ICCV}}, pages 1021--1030. {IEEE} Computer Society, 2017.

\bibitem[Carlini and Wagner(2017)]{DBLP:conf/sp/Carlini017}
Nicholas Carlini and David~A. Wagner.
\newblock Towards evaluating the robustness of neural networks.
\newblock In \emph{{IEEE} Symposium on Security and Privacy}, pages 39--57.
  {IEEE} Computer Society, 2017.

\bibitem[Cheng et~al.(2020)Cheng, Chen, and Chiu]{DBLP:conf/cvpr/ChengCC20}
Chia{-}Chi Cheng, Hung{-}Yu Chen, and Wei{-}Chen Chiu.
\newblock Time flies: Animating a still image with time-lapse video as
  reference.
\newblock In \emph{{CVPR}}, pages 5640--5649. {IEEE}, 2020.

\bibitem[Cozzolino et~al.(2018)Cozzolino, Thies, R{\"{o}}ssler, Riess,
  Nie{\ss}ner, and Verdoliva]{DBLP:journals/corr/abs-1812-02510}
Davide Cozzolino, Justus Thies, Andreas R{\"{o}}ssler, Christian Riess,
  Matthias Nie{\ss}ner, and Luisa Verdoliva.
\newblock Forensictransfer: Weakly-supervised domain adaptation for forgery
  detection.
\newblock \emph{CoRR}, abs/1812.02510, 2018.

\bibitem[Feng et~al.(2019)Feng, Cai, and Zhou]{DBLP:conf/nips/FengCZ19}
Ji~Feng, Qi{-}Zhi Cai, and Zhi{-}Hua Zhou.
\newblock Learning to confuse: Generating training time adversarial data with
  auto-encoder.
\newblock In \emph{NeurIPS}, pages 11971--11981, 2019.

\bibitem[Gondim{-}Ribeiro et~al.(2018)Gondim{-}Ribeiro, Tabacof, and
  Valle]{DBLP:journals/corr/abs-1806-04646}
George Gondim{-}Ribeiro, Pedro Tabacof, and Eduardo Valle.
\newblock Adversarial attacks on variational autoencoders.
\newblock \emph{CoRR}, abs/1806.04646, 2018.

\bibitem[Goodfellow et~al.(2015)Goodfellow, Shlens, and
  Szegedy]{DBLP:journals/corr/GoodfellowSS14}
Ian~J. Goodfellow, Jonathon Shlens, and Christian Szegedy.
\newblock Explaining and harnessing adversarial examples.
\newblock In \emph{{ICLR} (Poster)}, 2015.

\bibitem[Huang et~al.(2017)Huang, Wang, Luo, Ma, Jiang, Zhu, Li, and
  Liu]{DBLP:conf/cvpr/HuangWLMJZLL17}
Haozhi Huang, Hao Wang, Wenhan Luo, Lin Ma, Wenhao Jiang, Xiaolong Zhu, Zhifeng
  Li, and Wei Liu.
\newblock Real-time neural style transfer for videos.
\newblock In \emph{{CVPR}}, pages 7044--7052. {IEEE} Computer Society, 2017.

\bibitem[iperov(2020)]{DeepFaceLab-github}
iperov.
\newblock Deepfacelab github., 2020.
\newblock URL \url{https://github.com/iperov/DeepFaceLab}.

\bibitem[Khodabakhsh et~al.(2018)Khodabakhsh, Raghavendra, Raja, Wasnik, and
  Busch]{DBLP:conf/biosig/KhodabakhshRRWB18}
Ali Khodabakhsh, Ramachandra Raghavendra, Kiran~B. Raja, Pankaj~Shivdayal
  Wasnik, and Christoph Busch.
\newblock Fake face detection methods: Can they be generalized?
\newblock In \emph{{BIOSIG}}, volume {P-282} of \emph{{LNI}}, pages 1--6. {GI}
  / {IEEE}, 2018.

\bibitem[Kos et~al.(2018)Kos, Fischer, and Song]{DBLP:conf/sp/KosFS18}
Jernej Kos, Ian Fischer, and Dawn Song.
\newblock Adversarial examples for generative models.
\newblock In \emph{{IEEE} Symposium on Security and Privacy Workshops}, pages
  36--42. {IEEE} Computer Society, 2018.

\bibitem[Kurakin et~al.(2017)Kurakin, Goodfellow, and
  Bengio]{DBLP:conf/iclr/KurakinGB17a}
Alexey Kurakin, Ian~J. Goodfellow, and Samy Bengio.
\newblock Adversarial examples in the physical world.
\newblock In \emph{{ICLR} (Workshop)}. OpenReview.net, 2017.

\bibitem[Madry et~al.(2018)Madry, Makelov, Schmidt, Tsipras, and
  Vladu]{DBLP:conf/iclr/MadryMSTV18}
Aleksander Madry, Aleksandar Makelov, Ludwig Schmidt, Dimitris Tsipras, and
  Adrian Vladu.
\newblock Towards deep learning models resistant to adversarial attacks.
\newblock In \emph{{ICLR} (Poster)}. OpenReview.net, 2018.

\bibitem[Marra et~al.(2018)Marra, Gragnaniello, Cozzolino, and
  Verdoliva]{DBLP:conf/mipr/MarraGCV18}
Francesco Marra, Diego Gragnaniello, Davide Cozzolino, and Luisa Verdoliva.
\newblock Detection of gan-generated fake images over social networks.
\newblock In \emph{{MIPR}}, pages 384--389. {IEEE}, 2018.

\bibitem[Moosavi{-}Dezfooli et~al.(2017)Moosavi{-}Dezfooli, Fawzi, Fawzi, and
  Frossard]{DBLP:conf/cvpr/Moosavi-Dezfooli17}
Seyed{-}Mohsen Moosavi{-}Dezfooli, Alhussein Fawzi, Omar Fawzi, and Pascal
  Frossard.
\newblock Universal adversarial perturbations.
\newblock In \emph{{CVPR}}, pages 86--94. {IEEE} Computer Society, 2017.

\bibitem[Mu{\~{n}}oz{-}Gonz{\'{a}}lez et~al.(2017)Mu{\~{n}}oz{-}Gonz{\'{a}}lez,
  Biggio, Demontis, Paudice, Wongrassamee, Lupu, and
  Roli]{DBLP:conf/ccs/Munoz-GonzalezB17}
Luis Mu{\~{n}}oz{-}Gonz{\'{a}}lez, Battista Biggio, Ambra Demontis, Andrea
  Paudice, Vasin Wongrassamee, Emil~C. Lupu, and Fabio Roli.
\newblock Towards poisoning of deep learning algorithms with back-gradient
  optimization.
\newblock In \emph{AISec@CCS}, pages 27--38. {ACM}, 2017.

\bibitem[R{\"{o}}ssler et~al.(2019)R{\"{o}}ssler, Cozzolino, Verdoliva, Riess,
  Thies, and Nie{\ss}ner]{DBLP:conf/iccv/RosslerCVRTN19}
Andreas R{\"{o}}ssler, Davide Cozzolino, Luisa Verdoliva, Christian Riess,
  Justus Thies, and Matthias Nie{\ss}ner.
\newblock Faceforensics++: Learning to detect manipulated facial images.
\newblock In \emph{{ICCV}}, pages 1--11. {IEEE}, 2019.

\bibitem[Ruiz et~al.(2020)Ruiz, Bargal, and
  Sclaroff]{DBLP:journals/corr/abs-2003-01279}
Nataniel Ruiz, Sarah~Adel Bargal, and Stan Sclaroff.
\newblock Disrupting deepfakes: Adversarial attacks against conditional image
  translation networks and facial manipulation systems.
\newblock \emph{CoRR}, abs/2003.01279, 2020.

\bibitem[Szegedy et~al.(2014)Szegedy, Zaremba, Sutskever, Bruna, Erhan,
  Goodfellow, and Fergus]{DBLP:journals/corr/SzegedyZSBEGF13}
Christian Szegedy, Wojciech Zaremba, Ilya Sutskever, Joan Bruna, Dumitru Erhan,
  Ian~J. Goodfellow, and Rob Fergus.
\newblock Intriguing properties of neural networks.
\newblock In \emph{{ICLR} (Poster)}, 2014.

\bibitem[Tabacof et~al.(2016)Tabacof, Tavares, and
  Valle]{DBLP:journals/corr/TabacofTV16}
Pedro Tabacof, Julia Tavares, and Eduardo Valle.
\newblock Adversarial images for variational autoencoders.
\newblock \emph{CoRR}, abs/1612.00155, 2016.

\bibitem[Thies et~al.(2016)Thies, Zollh{\"{o}}fer, Stamminger, Theobalt, and
  Nie{\ss}ner]{DBLP:conf/cvpr/ThiesZSTN16}
Justus Thies, Michael Zollh{\"{o}}fer, Marc Stamminger, Christian Theobalt, and
  Matthias Nie{\ss}ner.
\newblock Face2face: Real-time face capture and reenactment of {RGB} videos.
\newblock In \emph{{CVPR}}, pages 2387--2395. {IEEE} Computer Society, 2016.

\bibitem[Toews(2020)]{forbes-deepfakes-bad}
Rob Toews.
\newblock Deepfakes are going to wreak havoc on society. we are not prepared.,
  2020.
\newblock URL
  \url{https://www.forbes.com/sites/robtoews/2020/05/25/deepfakes-are-going-to-wreak-havoc-on-society-we-are-not-prepared/\#2248f32b7494}.

\bibitem[Torzdf(2020)]{deepfakes-github}
Torzdf.
\newblock Deepfakes github., 2020.
\newblock URL \url{https://github.com/deepfakes/faceswap}.

\bibitem[Tram{\`{e}}r et~al.(2018)Tram{\`{e}}r, Kurakin, Papernot, Goodfellow,
  Boneh, and McDaniel]{DBLP:conf/iclr/TramerKPGBM18}
Florian Tram{\`{e}}r, Alexey Kurakin, Nicolas Papernot, Ian~J. Goodfellow, Dan
  Boneh, and Patrick~D. McDaniel.
\newblock Ensemble adversarial training: Attacks and defenses.
\newblock In \emph{{ICLR} (Poster)}. OpenReview.net, 2018.

\bibitem[Wang et~al.(2020{\natexlab{a}})Wang, Cho, and
  Yoon]{DBLP:journals/ral/WangCY20}
Lin Wang, Wonjune Cho, and Kuk{-}Jin Yoon.
\newblock Deceiving image-to-image translation networks for autonomous driving
  with adversarial perturbations.
\newblock \emph{{IEEE} Robotics Autom. Lett.}, 5\penalty0 (2):\penalty0
  1421--1428, 2020{\natexlab{a}}.

\bibitem[Wang et~al.(2020{\natexlab{b}})Wang, Juefei{-}Xu, Ma, Xie, Huang,
  Wang, and Liu]{DBLP:conf/ijcai/WangJMXHWL20}
Run Wang, Felix Juefei{-}Xu, Lei Ma, Xiaofei Xie, Yihao Huang, Jian Wang, and
  Yang Liu.
\newblock Fakespotter: {A} simple yet robust baseline for spotting
  ai-synthesized fake faces.
\newblock In \emph{{IJCAI}}, pages 3444--3451. ijcai.org, 2020{\natexlab{b}}.

\bibitem[Wang et~al.(2020{\natexlab{c}})Wang, Wang, Zhang, Owens, and
  Efros]{DBLP:conf/cvpr/WangW0OE20}
Sheng{-}Yu Wang, Oliver Wang, Richard Zhang, Andrew Owens, and Alexei~A. Efros.
\newblock Cnn-generated images are surprisingly easy to spot... for now.
\newblock In \emph{{CVPR}}, pages 8692--8701. {IEEE}, 2020{\natexlab{c}}.

\bibitem[Willetts et~al.(2019)Willetts, Camuto, Rainforth, Roberts, and
  Holmes]{willetts2019improving}
Matthew Willetts, Alexander Camuto, Tom Rainforth, Stephen Roberts, and Chris
  Holmes.
\newblock Improving vaes' robustness to adversarial attack.
\newblock \emph{arXiv preprint arXiv:1906.00230}, 2019.

\bibitem[Yeh et~al.(2020)Yeh, Chen, Tsai, and Wang]{DBLP:conf/wacv/YehCTW20}
Chin{-}Yuan Yeh, Hsi{-}Wen Chen, Shang{-}Lun Tsai, and Shang{-}De Wang.
\newblock Disrupting image-translation-based deepfake algorithms with
  adversarial attacks.
\newblock In \emph{{WACV} Workshops}, pages 53--62. {IEEE}, 2020.

\bibitem[Zhang et~al.(2017)Zhang, Zhu, Lei, Shi, Wang, and
  Li]{DBLP:journals/corr/abs-1708-05237}
Shifeng Zhang, Xiangyu Zhu, Zhen Lei, Hailin Shi, Xiaobo Wang, and Stan~Z. Li.
\newblock S\({}^{\mbox{3}}\)fd: Single shot scale-invariant face detector.
\newblock \emph{CoRR}, abs/1708.05237, 2017.

\bibitem[Zhou et~al.(2017)Zhou, Han, Morariu, and
  Davis]{DBLP:conf/cvpr/ZhouHMD17}
Peng Zhou, Xintong Han, Vlad~I. Morariu, and Larry~S. Davis.
\newblock Two-stream neural networks for tampered face detection.
\newblock In \emph{{CVPR} Workshops}, pages 1831--1839. {IEEE} Computer
  Society, 2017.

\end{thebibliography}

\end{document}